\documentclass{article}
\usepackage{spconf,amsmath,graphicx}
\usepackage[sort&compress]{natbib} 
\bibpunct{[}{]}{,}{n}{,}{,}
\usepackage{url}
\usepackage[ngerman, american]{babel}
\usepackage{booktabs}
\usepackage{amsmath}
\usepackage{color}
\usepackage{url}
\newcommand{\eg}{\textit{e.g.}}
\newcommand{\ie}{\textit{i.e.}}
\newcommand\norm[1]{\left\lVert#1\right\rVert}

\usepackage{cleveref}
\crefname{appendix}{App.\negthinspace\,}{App.\negthinspace\,}
\crefname{chapter}{Chap.\negthinspace\,}{Chap.\negthinspace\,}
\crefname{equation}{Eq.\negthinspace\,}{Eq.\negthinspace\,}
\crefname{algorithm}{Alg.\negthinspace\,}{Alg.\negthinspace\,}
\crefname{section}{Sec.\negthinspace\,}{Sec.\negthinspace\,}
\crefname{subsection}{Sec.\negthinspace\,}{Sec.\negthinspace\,}
\crefname{subsubsection}{Sec.\negthinspace\,}{Sec.\negthinspace\,}
\crefname{figure}{Fig.\negthinspace\,}{Fig.\negthinspace\,}
\crefname{table}{Tab.\negthinspace\,}{Tab.\negthinspace\,}
\crefname{subfigure}{Fig.\negthinspace\,}{Fig.\negthinspace\,}
\crefname{subsubfigure}{Fig.\negthinspace\,}{Fig.\negthinspace\,}
\crefname{lstlisting}{Lst.\negthinspace\,}{Lst.\negthinspace\,}

\title{Cell Segmentation in 3D Confocal Images using Supervoxel Merge-Forests with CNN-based Hypothesis Selection}

\name{{\parbox[c]{\textwidth}{\centering Johannes~Stegmaier$^{1,2,*}$ \qquad Thiago V. Spina$^{2,3}$ \qquad Alexandre X. Falc{\~{a}}o$^{4}$ \qquad Andreas Bartschat$^{1}$ \qquad Ralf~Mikut$^{1}$ \qquad Elliot~Meyerowitz$^{5}$ \qquad Alexandre~Cunha$^{2,*}$ \thanks{We are grateful for funding by the Helmholtz Association in the program BioInterfaces in Technology and Medicine
(RM), the German Research Foundation DFG in the project MI1315/4-1 (JS, RM), the Center for Advanced Methods in Biological Image Analysis at the Beckman Institute (JS, TS, EM, AC), the Howard Hughes Medical Institute (EM), the Gordon and Betty Moore Foundation (EM and AC) and the S{\~{a}}o Paulo Research Foundation in projects 2016/11853-2, 2015/09446-7, and 2014/12236-1 (TS, AF). The Titan Xp used for this research was donated by the NVIDIA Corporation. Correspondence: \texttt{johannes.stegmaier@kit.edu} or \texttt{cunha@caltech.edu}.}}}}

\address{
\small{$^{1}$Institute for Applied Computer Science, Karlsruhe Institute of Technology, Karlsruhe, Germany}\\
\small{$^{2}$Center for Advanced Methods in Biological Image Analysis, California Institute of Technology (Caltech), Pasadena, CA, USA}\\
\small{$^{3}$Brazilian Synchrotron Light Laboratory, CNPEM, Campinas, SP, Brazil}\\ 
\small{$^{4}$Institute of Computing, University of Campinas, Campinas, SP, Brazil}\\ 
\small{$^{5}$Howard Hughes Medical Institute and Division of Biology and Biological Engineering, Caltech, Pasadena, CA, USA}}

\begin{document}
%
\maketitle
\begin{abstract}Automated segmentation approaches are crucial to quantitatively analyze large-scale 3D microscopy images. Particularly in deep tissue regions, automatic methods still fail to provide error-free segmentations. To improve the segmentation quality throughout imaged samples, we present a new supervoxel-based 3D segmentation approach that outperforms current methods and reduces the manual correction effort. The algorithm consists of gentle preprocessing and a conservative super-voxel generation method followed by supervoxel agglomeration based on local signal properties and a postprocessing step to fix under-segmentation errors using a Convolutional Neural Network. We validate the functionality of the algorithm on manually labeled 3D confocal images of the plant \emph{Arabidopis thaliana} and compare the results to a state-of-the-art meristem segmentation algorithm.
\end{abstract}
\section{Introduction}
The shoot apical meristem present in flowering plants is comprised of a network of stem cells responsible for all above ground development of the plant. These tightly arranged cells partition the space into approximately convex polyhedra, resembling a Voronoi diagram, and forming a dome-like structure. To support scientists in their quest to understand the mechanisms that lead to cell development in the meristem and consequently plant growth, it is helpful to automatically quantify the geometry and topology of the network. With this information one can then, \eg, compute the spatial-temporal distribution of cell size from images obtained after lab experiments, use measurements to assist in the creation and validation of mathematical models of development and cell-cell communication, and carry on computational simulations on more faithful representations of the existing network. The work we present here is a step towards the goal of automatically quantifying the meristem of the {\it Arabidopsis thaliana} plant, with possible applications in similar images.

\begin{figure}[!ht]
 \includegraphics[width=\columnwidth]{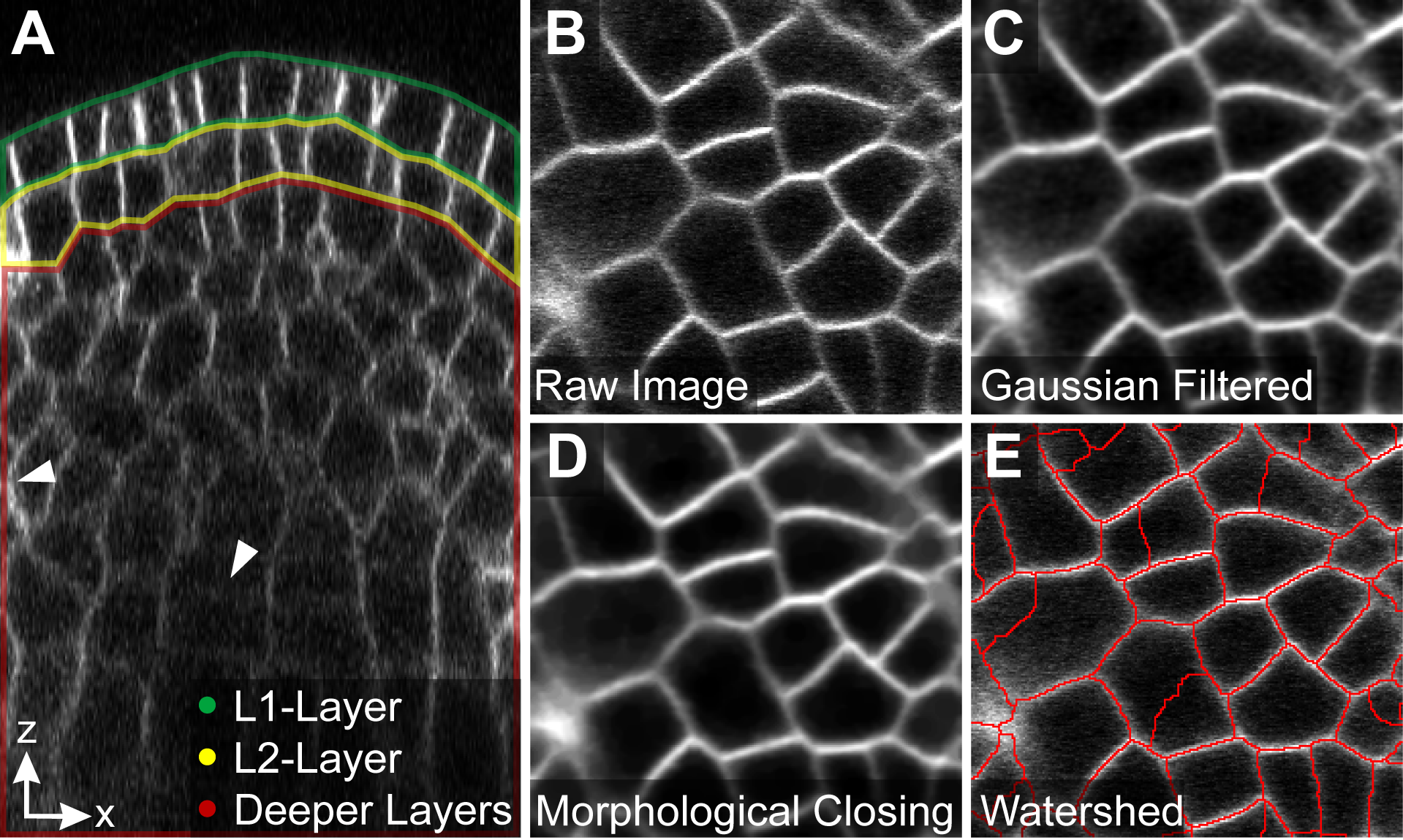}
 \vspace{-0.6cm}
\caption{(A) Cross-section through the meristem showing large intensity variations for deeper layers (lower arrow shows weak membrane signal). (B) Raw image in xy orientation, (C) Gaussian smoothed raw image, (D) iterative morphological closing, (E) watersheds of the initial over-segmentation in red superimposed on the raw image.}
\label{fig:Figure1}
\end{figure}

Several automated approaches have been presented in the past \cite{Fernandez10, Pop13, Reuille15, Stegmaier16} to quantitatively analyze potentially thousands of cells. Most existing methods preprocess the images to clean and boost the signal on cell plasma membranes, \eg, using mathematical morphology \cite{Fernandez10,Khan14a,Stegmaier16}, Hessian-based edge enhancement \cite{Mosaliganti12, Stegmaier16} or anisotropic filtering \cite{Pop13}, followed by the actual segmentation using a 3D watershed algorithm \cite{Fernandez10, Mosaliganti12, Khan14a}, a combinatorial fusion of 2D segmentations \cite{Stegmaier16} or variational methods \cite{Pop13, Faure16}. Finally, postprocessing heuristics are often used to suppress spurious detections based on volume constraints. 
While existing approaches work approximately well in the epidermis of the meristem, a region with better signal-to-noise ratio and comparable cell volumes, we found that most methods have particular difficulties in deeper layers of the meristem (\cref{fig:Figure1}A). Large variations of the fluorescent signal affect the performance of global, intensity-based parameters like $h$-minima that are frequently used for initialization of the 3D watershed and may cause leakage across weak boundaries or may erroneously split cells due to being set too sensitively. Hessian-based plane enhancement frequently fails to properly reconstruct the signal of en-face membranes and may even remove faint edges, resulting in under-segmentation errors in the axial direction. 

To overcome these limitations and to improve the quality of automatic segmentation also in deeper tissue layers, we largely skip sophisticated preprocessing steps that potentially introduce additional artifacts. Instead, we try to obtain an initial over-segmentation that partitions the image using a set of supervoxels that does not span across cell boundaries. We then agglomerate neighboring supervoxels based on prior knowledge to form a set of merge-trees (\eg, based on edge intensity, object volume and shape). As a final postprocessing step, under-segmentation errors that were caused by erroneous fusion of supervoxels are resolved using a Convolutional Neural Network (CNN). The algorithm does not rely on a nucleus channel to provide seeds for the segmentation and solely uses features present in the membrane images. Moreover, a key advantage of the algorithm is the need of few parameters making it easy to adjust and apply to new image data. We validated the performance of the proposed algorithm on manually annotated 3D confocal microscopy images of the \emph{Arabidopis thaliana} shoot apical meristem and compare the results to existing cell shape segmentation algorithms \cite{Fernandez10, Mosaliganti12}.

\section{Supervoxel-based 3D Segmentation}
Supervoxel-based segmentation approaches have been successfully applied in various computer vision domains (including segmentation of 2D and 3D microscopy images) and form the basis of the method presented in this contribution \cite{Funke15, Stegmaier16, Liu16b, Wassenberg09}. The approaches consist of mainly three steps: (1) a preprocessing stage to improve the image signal and to enhance the desired object boundaries, (2) a supervoxel generation step that partitions the enhanced image into meaningful parts and (3) a supervoxel merging phase that tries to agglomerate existing supervoxels to the desired objects of interest. An important precondition that has to be met by the supervoxel algorithm is that identified regions should not span over multiple objects as these under-segmentation errors cannot be corrected in subsequent merging steps.

\subsection{Preprocessing and Supervoxel Generation}
Starting with the raw input image (\cref{fig:Figure1}B), we use a 3D Gaussian filter to reduce the high-frequency noise in the image (\cref{fig:Figure1}C). Subsequently, we apply an iterative morphological closing to the Gaussian-smoothed image, in order to close membrane gaps \cite{Vachier05, Stegmaier16} (\cref{fig:Figure1}D). The radius for the iterative closing should be selected such that the maximum radius is smaller than the radius of the smallest objects that need to be resolved. Throughout the experiments performed for this paper, we used a maximum closing radius of $R^\text{cl}_\text{max} = 3$.

For supervoxel generation we apply a morphological watershed algorithm that is initialized with all local minima present in the preprocessed image (\cref{fig:Figure1}E) \cite{Beare06}. By skipping the frequently used \emph{h}-minima transform in the first place, we strongly reduce under-segmentation errors and the need to manually optimize this parameter locally, as done in \cite{Willis16}. However, using all local minima potentially generates several split segments per cell that need to be combined to complete cells in a subsequent supervoxel merging phase using a region adjacency graph.

\subsection{Merge-Forest Generation of 3D Supervoxels}
The following step is the generation of a merge-forest, \ie, adjacent supervoxels are iteratively merged on the basis of volume- and intensity-based edge features to a set of merge-trees. Each node of a merge-tree corresponds to a potential segmentation hypothesis. In contrast to existing methods that precompute a single merge-tree for the entire image \cite{Funke15, Liu16b, Couprie05}, we terminate the merging phase based on a specimen-dependent maximum volume constraint. This allows us to pick the best hypotheses using a CNN-based postprocessing step that only needs to be trained on small image snippets that maximally cover a few cells rather than all levels of detail of the entire image.

\subsubsection{Minimum Volume Condition}
\label{sec:MinimumVolumeCriterion}
The first merge feature uses the minimum volume $V_{\text{min}}$ to determine which of the supervoxels is smaller than the smallest cell and needs to be merged. This specimen-dependent parameter can be approximated, \eg, by measuring the smallest radius $R_{\text{min}}$ of any object within the data set. By approximating the cells as spherical objects, we calculate $V_{\text{min}} = \frac{4}{3}\cdot\pi\cdot R_{\text{min}}^3$. If the segmented objects substantially violate the sphericity assumption, $V_{\text{min}}$ can be replaced by measurements of manual segmentations or textbook knowledge. To identify which of the supervoxels should be merged, we linearly map all values between zero and $V_{\text{min}}$ to the interval $[0,1]$:
\begin{equation}
	f_{V_{\text{min}}}(i, j) = \max\left(0, \min\left(1, \frac{\min{(V_i, V_j)}}{V_{\text{min}}}\right)\right).
\end{equation}

If the volume of the smaller merging candidate is below $V_{\text{min}}$, the supervoxels $i$ and $j$ need to be merged. To ensure that edges with smaller intensities are collapsed first \cite{Funke15}, we multiply $f_{V_{\text{min}}}(i, j)$ with the average intensity $\mu^{\text{cmn}}_{ij}$ of the shared boundary between the two segments $i$ and $j$ if $f_{V_{\text{min}}}(i, j) < 1$. If the volume of segment $i$ or $j$ is larger than $V_{\text{min}}$, $f_{V_{\text{min}}}$ evaluates to $1$ and does not trigger further merges.

\subsubsection{Bright Homogeneous Boundary Condition}
\label{sec:BoundaryCondition}
\begin{figure}[!ht]
 \includegraphics[width=\columnwidth]{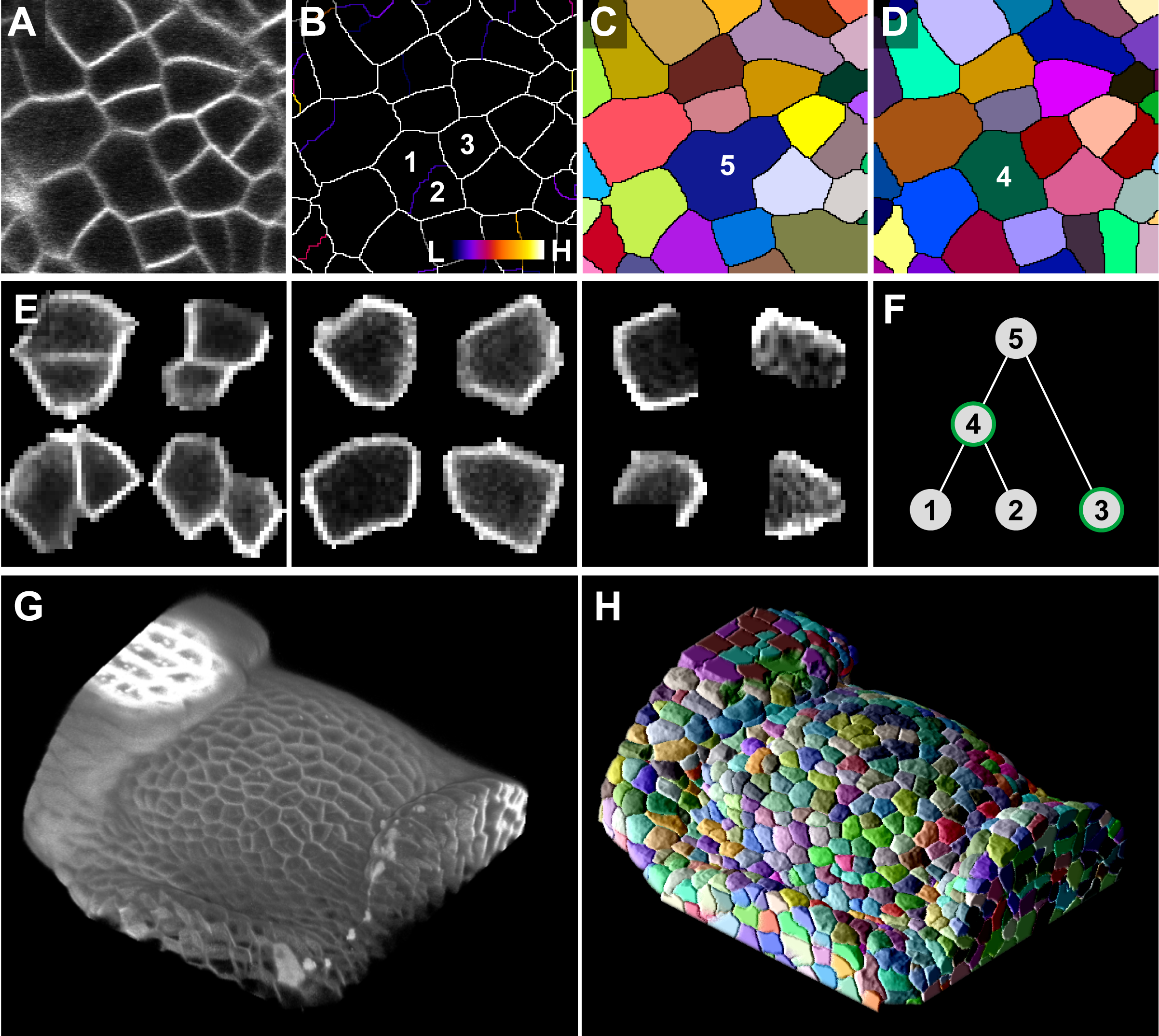}
 \vspace{-0.6cm}
\caption{(A) Raw image (see \cref{fig:Figure1}B), (B) initial over-segmentation using the sort feature for coloring, (C) segments after the supervoxel merging, (D) final segmentation with CNN-based correction of under-segmentation errors. (E) Example images for the classes \emph{under-segmentation}, \emph{correct cell} and \emph{over-segmentation} that were used for CNN training. (F) Merge-tree for the supervoxels indicated in (B-D). (G,H) 3D rendition of the raw image and the final segmentation.}
\label{fig:Figure2}
\end{figure}

The second merge feature is based on the assumption that each cell should be surrounded by a bright boundary with a darker interior where no fluorescent marker is expressed (\cref{fig:Figure2}A). We compare the intensity difference of the mean intensity on the shared boundary $\mu^{\text{cmn}}_{ij}$ to the mean intensity of the entire cell including the boundaries $\mu^{\text{total}}_{ij}$ versus the intensity on the boundary after fusion of the segments $\mu^{\text{bdry}}_{ij}$:
\begin{equation}
	f_{\text{bdry}}(i,j) = \min \left(1, \frac{\left|\mu^{\text{cmn}}_{ij} - \mu^{\text{total}}_{ij}\right|}{\left|\mu^{\text{cmn}}_{ij} - \mu^{\text{bdry}}_{ij}\right|} \right).
\end{equation}

A value of $f_{\text{bdry}}(i,j)$ smaller than $1$ indicates that the intensity difference of the considered intersection to the intensity of the entire cell including the boundary is smaller than the intensity of the boundary of the fused cell. Thus, the shared boundary is likely to be a part of the cell interior and may be removed. Instead of using absolute intensity values, the measure is based on relative intensity differences to avoid that region merges in low-intensity regions are more likely than region merges in high-contrast regions. 

\subsubsection{Edge Feature-based Merge-Forest Generation}
\label{sec:MergeForestGeneration}
In order to merge supervoxels using multiple features, all features need to be normalized such that none of the features suppresses the others due to a larger value range. All employed edge features lie in the interval $[0,1]$, with $1$ indicating that no merge should be performed and a value closer to zero indicating that a merge is likely (\cref{fig:Figure2}B). All features are then combined to a feature vector, which is divided by the square root of the number of features for normalization:
\begin{equation}
	f_{\text{sort}}(i,j) = \left(\sqrt{2}\right)^{-1} \cdot \norm{ \left( \begin{array}{c} f_{V_{\text{min}}}(i, j) \\ f_{\text{bdry}}(i,j) \end{array} \right)}.
\label{eq:FeatureCombination}
\end{equation}

As long as $f_{\text{sort}}(i,j) < 1$, at least one of the constraints for a valid segment boundary is violated. For each pair of adjacent supervoxels $i$ and $j$, we compute $f_{\text{sort}}(i,j)$ and sort the merge queue in ascending order based on this feature. The algorithm sequentially processes the merge queue, by merging the two supervoxels connected by the edge with the lowest merge score (\cref{fig:Figure2}C). After a merge has been performed, all edges previously connected to one of the two merged supervoxels are updated and inserted to the merge queue at the appropriate location. Using optimized data structures that keep track of the connectivity, intensity properties and merge hierarchy of all supervoxels, merge features can be recomputed recursively. This provides a fast way of updating edge features after merging without further iterations over all voxels in subsequent steps. Additionally, each performed merge is memorized and stored as a merge-tree, yielding a merge-forest when merging supervoxels is no longer valid.  To prevent the merge-trees from growing too large, edges are skipped if the combination of the two neighboring supervoxels would violate the maximum volume criterion. 

\begin{figure*}[!ht]
 \includegraphics[width=\textwidth]{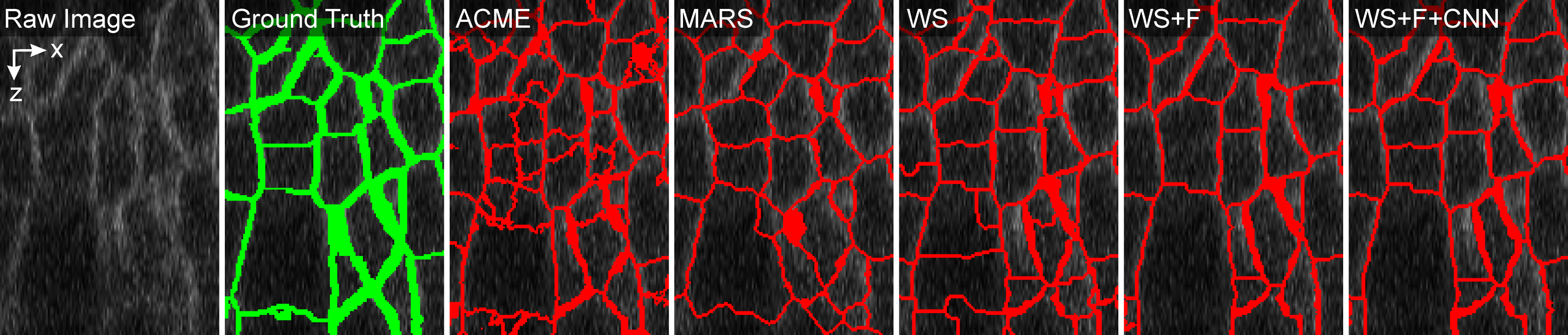}
 \vspace{-0.6cm}
\caption{An axial cross-section of the raw image, the manually annotated ground truth and respective segmentation boundaries in red from different methods (left to right).}
\label{fig:Figure3}
\end{figure*}
\subsection{CNN-based Selection of Merge-Tree Nodes}
As a final post-processing step, we use a Convolutional Neural Network to classify all hypotheses of all merge-trees and to resolve under-segmentations that are potentially introduced by the region merging step (\cref{fig:Figure2}D). The classification is based on 3D image patches of size $32\times 32 \times 32$ voxels that contain either a single correct cell, a cell segment (over-segmentation) or multiple connected cells (under-segmentation). The data for training and testing the network was extracted from the top layers of 20 data sets of the shoot apical meristem in \emph{A. thaliana} (\cref{fig:Figure2}E) \cite{Willis16}. 
We adapt TensorFlow's Deep MNIST network to work with 3D input images using two convolution layers ($5\times 5 \times 5$), each followed by a max pooling layer ($2\times 2 \times 2$, stride 2). These layers yield tensors with 32 and 64 features, respectively. Finally, we use a fully connected layer with 1024 output nodes with a dropout probability of $0.5$ that is connected to a 3-class softmax layer.
Training was performed using the ADAM optimizer on the cross-entropy loss function implemented in Google's TensorFlow API and executed on an NVIDIA Titan Xp. To use the identified class probabilities to correct under-segmentation errors, all merge-trees are revisited in a top-down fashion. If the under-segmentation probability is larger than one of the other classes, the current segmentation hypothesis (\ie, the root node of the current merge-tree) is discarded and the hypotheses of the daughter nodes are checked instead. This process is repeated recursively until one of the probabilities of the other classes exceeds the under-segmentation probability or upon reaching a leaf node of the merge-tree (\cref{fig:Figure2}F).

\section{Segmentation of the Shoot Apical Meristem in A. Thaliana}
We first applied the new pipeline to $124$ 3D confocal image stacks of \emph{A. thaliana} with manually corrected L1 and L2 layers \cite{Willis16}. On average, the proposed algorithm achieved a precision of $0.991\pm0.008$, a recall of $0.973\pm0.012$ and an F-score of $0.982\pm0.008$ for the L1 layer and a precision of $0.960\pm0.033$, a recall of $0.932\pm0.023$ and an F-score of $0.945\pm0.023$ for the L2 layer. Due to the limited batch-processing capabilities of the current implementation, the CNN-based postprocessing was not yet enabled for these experiments and we expect that the recall will further improve upon adding this under-segmentation correction step.

To investigate the segmentation quality in deeper layers as well, we densely labeled a $128 \times 128 \times 200$ image region using SEGMENT3D, a new interactive and collaborative 3D segmentation correction tool \cite{Spina18}. \cref{fig:Figure3} and \cref{tab:SegmentationPerformance} compare the segmentation performance of ACME \cite{Mosaliganti12}, MARS \cite{Fernandez10} and three intermediate steps of the proposed algorithm both qualitatively and quantitatively. The final pipeline including the CNN-based correction module reaches the highest F-Score of $0.943$ and successfully corrected under-segmentation errors that were introduced by the merging step, yielding an improvement of $0.028$ compared to the best existing approach. 
\begin{table}
\caption{3D Segmentation Performance in Deeper Layers}
\resizebox{\columnwidth}{!}{
\begin{tabular}{lccc}
\toprule
\textbf{Algorithm} 		& \textbf{Precision} & \textbf{Recall} & \textbf{F-Score} \\
\midrule
ACME					& 0.745 & \textbf{0.976} & 0.845 \\
MARS					& 0.921 & 0.909 & 0.915 \\
Watershed				& 0.844 & 0.947 & 0.893 \\
Watershed+Fusion		& \textbf{0.982} & 0.881 & 0.929 \\
Watershed+Fusion+CNN	& 0.957 & 0.929 & \textbf{0.943} \\
\bottomrule
\end{tabular}}
\label{tab:SegmentationPerformance}
\end{table}
\vspace{-0.75cm}
\section{Conclusion}
In this contribution, we present a new supervoxel-based approach to 3D cell shape reconstruction comprising an initial over-segmentation, an iterative supervoxel merging step and a CNN-based under-segmentation correction. The algorithm requires only two parameters, the minimum and maximum volume of cells. Even in deeper tissue layers that exhibit varying signal intensities and weak en-face membranes, the approach proved to reliably reconstruct 3D cell shapes for the most part. We validate the algorithm on manually labeled 3D confocal microscopy images of \emph{A. thaliana} and quantify the improvements compared to state-of-the-art algorithms for cell shape reconstruction \cite{Fernandez10, Mosaliganti12}. An initial version of the pipeline is implemented in the open-source software XPIWIT \cite{Bartschat16}.

To improve the generalization capabilities of the CNN and to ideally make it work across different data sets and across specimens, we plan to train the network on both manually annotated and synthetic 3D microscopy data that represent and mimic various kinds of imaging circumstances \cite{Stegmaier16a, Weigert17}. In the current implementation, the preprocessing, the region merging and the over-segmentation correction modules are implemented in C++, MATLAB and Python. In future releases, we plan to streamline the pipeline, such that the entire approach can be directly performed in a single XPIWIT pipeline \cite{Bartschat16}. Furthermore, we plan to replace ITK's serial 3D watershed implementation by a parallel watershed implementation, to eradicate the current bottleneck of the pipeline and to make the algorithm applicable to terabyte-scale 3D+t data sets.

\bibliographystyle{diss}
\bibliography{Bibliography}

\begin{thebibliography}{10}

\bibitem{Bartschat16}
\textsc{Bartschat, A.}; \textsc{H{\"u}bner, E.}; \textsc{Reischl, M.};
  \textsc{Mikut, R.}; \textsc{Stegmaier, J.}: {XPIWIT} - An {XML} Pipeline
  Wrapper for the Insight Toolkit. \emph{Bioinformatics} 32 (2016) 2,
  p.~315--317.

\bibitem{Beare06}
\textsc{Beare, R.}; \textsc{Lehmann, G.}: The Watershed Transform in
  ITK-Discussion and New Developments. \emph{The Insight Journal}  (2006),
  p.~1--24.

\bibitem{Couprie05}
\textsc{Couprie, M.}; \textsc{Najman, L.}; \textsc{Bertrand, G.}: Quasi-Linear
  Algorithms for the Topological Watershed. \emph{Journal of Mathematical
  Imaging and Vision} 22 (2005) 2-3, p.~231--249.

\bibitem{Faure16}
\textsc{Faure, E.}; \textsc{Savy, T.}; \textsc{Rizzi, B.}; \textsc{Melani, C.};
  \textsc{Sta{\v{s}}ov{\'a}, O.}; \textsc{Fabr{\`e}ges, D.};
  \textsc{{\v{S}}pir, R.}; \textsc{Hammons, M.};
  \textsc{{\v{C}}{\'u}nderl{\'\i}k, R.}; \textsc{Recher, G.}; \textsc{et~al.}:
  A Workflow to Process {3D}+Time Microscopy Images of Developing Organisms and
  Reconstruct Their Cell Lineage. \emph{Nature Communications} 7 (2016) 8674,
  p.~1--10.

\bibitem{Fernandez10}
\textsc{Fernandez, R.}; \textsc{Das, P.}; \textsc{Mirabet, V.};
  \textsc{Moscardi, E.}; \textsc{Traas, J.}; \textsc{Verdeil, J.-L.};
  \textsc{Malandain, G.}; \textsc{Godin, C.}: Imaging Plant Growth in {4D}:
  Robust Tissue Reconstruction and Lineaging at Cell Resolution. \emph{Nature
  Methods} 7 (2010) 7, p.~547--553.

\bibitem{Funke15}
\textsc{Funke, J.}; \textsc{Hamprecht, F.~A.}; \textsc{Zhang, C.}: Learning to
  Segment: Training Hierarchical Segmentation Under a Topological Loss. In:
  \emph{International Conference on Medical Image Computing and
  Computer-Assisted Intervention}, p. 268--275, Springer, 2015.

\bibitem{Khan14a}
\textsc{Khan, Z.}; \textsc{Wang, Y.-C.}; \textsc{Wieschaus, E.~F.};
  \textsc{Kaschube, M.}: Quantitative {4D} Analyses of Epithelial Folding
  during Drosophila Gastrulation. \emph{Development} 141 (2014) 14,
  p.~2895--2900.

\bibitem{Liu16b}
\textsc{Liu, T.}; \textsc{Seyedhosseini, M.}; \textsc{Tasdizen, T.}: Image
  Segmentation using Hierarchical Merge Tree. \emph{IEEE Transactions on Image
  Processing} 25 (2016) 10, p.~4596--4607.

\bibitem{Mosaliganti12}
\textsc{Mosaliganti, K.~R.}; \textsc{Noche, R.~R.}; \textsc{Xiong, F.};
  \textsc{Swinburne, I.~A.}; \textsc{Megason, S.~G.}: ACME: Automated Cell
  Morphology Extractor for Comprehensive Reconstruction of Cell Membranes.
  \emph{PLoS Computational Biology} 8 (2012) 12, p.~e1002780.

\bibitem{Pop13}
\textsc{Pop, S.}; \textsc{Dufour, A.~C.}; \textsc{Le~Garrec, J.-F.};
  \textsc{Ragni, C.~V.}; \textsc{Cimper, C.}; \textsc{Meilhac, S.~M.};
  \textsc{Olivo-Marin, J.-C.}: Extracting 3D Cell Parameters from Dense Tissue
  Environments: Application to the Development of the Mouse Heart.
  \emph{Bioinformatics} 29 (2013) 6, p.~772--779.

\bibitem{Reuille15}
\textsc{de~Reuille, P.~B.}; \textsc{Routier-Kierzkowska, A.-L.};
  \textsc{Kierzkowski, D.}; \textsc{Bassel, G.~W.}; \textsc{Sch{\"u}pbach, T.};
  \textsc{Tauriello, G.}; \textsc{Bajpai, N.}; \textsc{Strauss, S.};
  \textsc{Weber, A.}; \textsc{Kiss, A.}; \textsc{et~al.}: MorphoGraphX: A
  Platform for Quantifying Morphogenesis in 4D. \emph{Elife} 4 (2015),
  p.~e05864.

\bibitem{Spina18}
\textsc{Spina, T.~V.}; \textsc{Stegmaier, J.}; \textsc{Falc{\~{a}}o, A.~X.};
  \textsc{Meyerowitz, E.}; \textsc{Cunha, A.}: SEGMENT3D: A Web-Based Tool for
  Collaborative Segmentation of 3D Microscopy Images of the Shoot Apical
  Meristem. In: \emph{Submitted for ISBI 2018}, 2018.

\bibitem{Stegmaier16}
\textsc{Stegmaier, J.}; \textsc{Amat, F.}; \textsc{Lemon, W.~B.};
  \textsc{McDole, K.}; \textsc{Wan, Y.}; \textsc{Teodoro, G.}; \textsc{Mikut,
  R.}; \textsc{Keller, P.~J.}: Real-Time Three-Dimensional Cell Segmentation in
  Large-Scale Microscopy Data of Developing Embryos. \emph{Developmental Cell}
  36 (2016) 2, p.~225--240.

\bibitem{Stegmaier16a}
\textsc{Stegmaier, J.}; \textsc{Arz, J.}; \textsc{Schott, B.}; \textsc{Otte,
  J.~C.}; \textsc{Kobitski, A.}; \textsc{Nienhaus, G.~U.}; \textsc{Str\"ahle,
  U.}; \textsc{Sanders, P.}; \textsc{Mikut, R.}: Generating Semi-Synthetic
  Validation Benchmarks for Embryomics. In: \emph{Proc., IEEE International
  Symposium on Biomedical Imaging: From Nano to Macro}, 2016.

\bibitem{Vachier05}
\textsc{Vachier, C.}; \textsc{Meyer, F.}: The Viscous Watershed Transform.
  \emph{Journal of Mathematical Imaging and Vision} 22 (2005) 2, p.~251--267.

\bibitem{Wassenberg09}
\textsc{Wassenberg, J.}; \textsc{Middelmann, W.}; \textsc{Sanders, P.}: An
  Efficient Parallel Algorithm for Graph-based Image Segmentation. In:
  \emph{International Conference on Computer Analysis of Images and Patterns},
  p. 1003--1010, Springer, 2009.

\bibitem{Weigert17}
\textsc{Weigert, M.}; \textsc{Royer, L.}; \textsc{Jug, F.}; \textsc{Myers, G.}:
  Isotropic Reconstruction of 3D Fluorescence Microscopy Images using
  Convolutional Neural Networks. \emph{arXiv preprint arXiv:1704.01510}
  (2017).

\bibitem{Willis16}
\textsc{Willis, L.}; \textsc{Refahi, Y.}; \textsc{Wightman, R.};
  \textsc{Landrein, B.}; \textsc{Teles, J.}; \textsc{Huang, K.~C.};
  \textsc{Meyerowitz, E.~M.}; \textsc{J{\"o}nsson, H.}: Cell Size and Growth
  Regulation in the Arabidopsis Thaliana Apical Stem Cell Niche.
  \emph{Proceedings of the National Academy of Sciences} 113 (2016) 51,
  p.~E8238--E8246.

\end{thebibliography}

\end{document}